# Index Light, Reason Deep: Deferred Visual Ingestion for Visual-Dense Document Question Answering

Tao Xu
OpsMate AI, Inc.
sxunix@sjtu.edu.cn, sxunix@gmail.com

## Abstract

Existing multimodal document question answering methods predominantly adopt a Pre-Ingestion (PI) strategy: during the indexing phase, a Vision Language Model (VLM) is called on every page to generate page descriptions that are then encoded into vectors, and questions are answered via embedding similarity retrieval. However, this approach faces a dual dilemma on visual-dense engineering documents: VLM blind descriptions inevitably lose critical visual details, and embedding retrieval systematically fails on highly similar documents.

This paper proposes the **Deferred Visual Ingestion (DVI)** framework: zero VLM calls during preprocessing, leveraging only document structural information (table of contents, drawing numbers) to automatically build a hierarchical index through the HDNC (Hierarchical Drawing Number Clustering) algorithm; during inference, candidate pages are located via BM25 retrieval, and the original images along with the specific question are sent to a VLM for targeted analysis.

Large-scale experiments on three datasets validate the effectiveness of DVI: on Bridge engineering drawings (1,323 questions), end-to-end QA accuracy reaches 65.6% vs. PI's 24.3% (+41.3pp); on Steel catalog (186 questions), 30.6% vs. 16.1% (+14.5pp); on CircuitVQA, a public benchmark (9,315 questions), retrieval ImgR@3 achieves 31.2% vs. 0.7%. On the Bridge dataset, we evaluated ColPali (ICLR 2025 visual retrieval SOTA), which achieved only 20.1% PageR@3, demonstrating that the failure of embedding retrieval on homogeneous engineering documents is structural rather than due to insufficient model capability. Ablation studies show that HDNC zero-cost automatic indexing yields a +27.5pp retrieval improvement, and VLM conversion rate analysis confirms that the bottleneck lies on the retrieval side rather than the comprehension side.

# 1. Introduction

## 1.1 The Problem: Challenges of Visual-Dense Document RAG

Visual-dense documents—engineering drawings, electrical schematics, product catalogs, architectural blueprints—are core knowledge carriers in industrial domains. The critical information in such documents exists in visual form (dimension annotations, pipeline routing, tabular data, component layouts), which traditional OCR pipelines can hardly process effectively. OHRBench [1] has demonstrated that OCR noise cascades through retrieval-augmented generation (RAG) pipelines, degrading performance from the retrieval layer all the way to the generation layer.

Current mainstream multimodal document RAG approaches predominantly adopt a **Pre-Ingestion** (PI) strategy: during the indexing phase, a Vision Language Model (VLM) is called on every page to generate page descriptions, which are encoded into embedding vectors and stored in a vector database; at query time, relevant pages are retrieved via vector similarity. This paradigm has achieved notable progress on general documents (VisRAG [2], ColPali [3], M3DocRAG [4]), but faces fundamental difficulties on visual-dense engineering documents.

## 1.2 The Dilemma of Pre-Ingestion

Pre-Ingestion approaches face two modes of failure on visual-dense documents:

**Information loss risk.** During preprocessing, the VLM has no knowledge of what the user will ask, and can only generate generic "blind descriptions." The critical information on engineering drawings—dimension annotations, terminal numbers, rebar specifications, pipeline connections—is highly dense and interrelated, making it inevitable that blind descriptions omit substantial visual details. Prior work [5] found that LLM summary preprocessing causes a 13% drop in retrieval precision, and the visual information density of engineering drawings far exceeds that of ordinary documents.

**Retrieval failure risk.** Even if blind descriptions successfully capture key information, embedding retrieval still systematically fails on highly similar documents. Drawing sets from large engineering projects contain hundreds of structurally similar drawings—for a single bridge project, each of 20 bridges has general arrangement drawings, reinforcement drawings, detail drawings, and so on. The VLM blind descriptions produce highly similar text, and the resulting vectors cluster tightly in the embedding space, where cosine similarity cannot effectively discriminate among them. This problem has received little attention in prior literature, as existing benchmarks mostly target heterogeneous document collections (e.g., Wikipedia pages) rather than homogeneous engineering document sets.

More critically, these two risks are **compounding and irrecoverable**: once a blind description omits information or embedding retrieval fails to hit the correct page, the system cannot go back and examine the original drawing at query time. The errors of Pre-Ingestion are unidirectional and irreversible.

## 1.3 The Core Idea of DVI: Index for Locating, Not Understanding

The root cause of the above dilemma lies in the fact that existing approaches attempt to "understand" every page during preprocessing, at a time when the user's actual concerns are unknown. This paper proposes the **Deferred Visual Ingestion (DVI)** framework, which adopts a fundamentally different strategy:

- **Preprocessing phase:** No VLM is called. Only the document's structural information (table of contents pages, drawing numbers, titles) is used to build a lightweight index, addressing the "**which page**" localization problem.
- **Inference phase:** After the user poses a question, the index locates candidate pages, and the original images along with the specific question are sent to a VLM for targeted analysis, addressing the "**what is it**" comprehension problem.

The design principle of DVI is "**Index for locating, not understanding**." This is analogous to lazy evaluation in computer science: rather than precomputing all possible descriptions, the system analyzes on demand only when first needed.

This simple idea yields powerful practical results. Large-scale experiments on three datasets show that DVI, with zero VLM preprocessing cost, comprehensively outperforms PI: end-to-end QA accuracy is higher by +41.3pp (Bridge engineering drawings) and +14.5pp (Steel catalog), and retrieval performance leads by over +30pp across all three datasets.

## 1.4 Contributions

1. **We propose the Deferred Visual Ingestion (DVI) framework.** Visual comprehension is deferred from the preprocessing phase to the inference phase, achieving substantially better end-to-end performance than Pre-Ingestion at zero VLM preprocessing cost. The framework is validated on 1,509 industrial questions (Bridge 1,323 + Steel 186) and 9,315 public benchmark questions (CircuitVQA).

2. **We propose the HDNC automatic indexing algorithm.** By exploiting the inherent structure of engineering drawing numbering systems, HDNC automatically discovers hierarchical categories and builds an index with zero API calls. On the Bridge dataset, HDNC automatic indexing yields a +27.5pp retrieval improvement, with recall even surpassing that of manually curated metadata extracted by VLMs.

3. **We reveal a text-quality-adaptive indexing strategy.** We discover that text fusion produces diametrically opposite effects across document types (vector PDF +21.3pp vs. scanned document −40.9pp), and propose an adaptive framework that dynamically adjusts the indexing strategy based on document type.

4. **We provide the first large-scale comparative experiment for engineering document RAG.** Across two types of industrial documents (vector PDF drawings + scanned document catalogs) and one public benchmark, we systematically compare DVI, PI, and ColPali (ICLR 2025 visual retrieval SOTA) at both the retrieval layer and the end-to-end QA layer. We find that ColPali achieves only 20.1% PageR@3 on highly homogeneous engineering drawings, even lower than naive PI-embedding (30.7%), demonstrating that the failure of embedding retrieval on such documents is structural in nature.

## 2. Related Work

### 2.1 Multimodal Retrieval-Augmented Generation

Retrieval-Augmented Generation (RAG) [6] has become the dominant paradigm for knowledge-intensive question answering. In recent years, RAG has expanded from pure text to multimodal documents [7]. UNIDOC-BENCH [8] demonstrated on 70K real PDF pages that multimodal fusion RAG consistently outperforms unimodal approaches. VisRAG [2] constructed a purely visual RAG pipeline (ICLR 2025), M3DocRAG [4] employed ColPali indexing combined with VLM-based visual question answering to achieve multi-page, multi-document understanding, ViDoRAG [9] proposed multi-agent iterative reasoning (EMNLP 2025), and VisDoMRAG [10] runs visual and textual RAG pipelines in parallel (NAACL 2025). These works share the design of "letting the VLM see the original image" at query time, but **all rely on heavyweight visual embedding models at the indexing stage**. DVI uses no visual model whatsoever at the indexing stage, relying solely on rule-based extraction of textual metadata.

### 2.2 Visual Document Retrieval

ColPali [3] introduced the paradigm of using vision-language models to directly generate visual embeddings for document pages. Subsequent work such as ColFlor [11] and ModernVBERT [12] continued to optimize model efficiency, while DSE [13] encodes document screenshots into dense vectors. These methods have demonstrated that visual embeddings can replace OCR for high-quality document retrieval, yet **the indexing stage still requires running vision models with billions of parameters**. More importantly, existing benchmarks are predominantly designed for heterogeneous document collections (e.g., Wikipedia, academic papers), and performance on homogeneous engineering drawing collections has not been evaluated. Our experiments show that embedding-based retrieval systematically fails on highly similar engineering documents.

## 2.3 Engineering Document Analysis

Research combining engineering drawings with RAG remains in its early stages. DesignQA [14] (ASME JCISE 2025) constructed an engineering document QA benchmark based on Formula SAE data and found that naive RAG performs poorly on engineering documents but proposed no architectural improvements. Blueprint [15] (2025) implemented retrieval functionality for engineering drawings but did not address QA. Work [16] found that VLMs excel at high-level reasoning but remain deficient in low-level visual perception (e.g., recognizing terminal numbers or IP addresses). Effective architectural design for engineering drawing RAG—particularly how to leverage the structured numbering systems of engineering documents for efficient retrieval—is virtually absent from the existing literature.

## 2.4 OCR Noise and Pre-Ingestion Information Loss

OHRBench [1] (ICCV 2025) demonstrated that no OCR solution can build a high-quality knowledge base for RAG—OCR noise cascades from the retrieval stage through to the generation stage. Lost in OCR Translation [17] found that visual methods are approximately 12% more robust to document quality degradation than OCR-based methods. Work [5] found that LLM summarization preprocessing leads to a 13% drop in retrieval accuracy. These findings support the design motivation of DVI's "preserve the original image, defer analysis" approach—intermediate representations produced by Pre-Ingestion (blind descriptions, OCR text, embedding vectors) inevitably introduce information loss, whereas DVI sends the original image directly to the VLM, eliminating intermediate failure points.

## 2.5 Positioning and Summary

Table 0 compares the key differences between DVI and representative prior work.

**Table 0: Key Differences Between DVI and Existing Methods**

| Dimension | ColPali Family | VisRAG/M3DocRAG | **DVI (Ours)** |
|---|---|---|---|
| VLM at Indexing | Required (3B–4B) | Required | **Not required** |
| VLM at Query Time | Not applicable | Required | On-demand |
| Retrieval Method | Visual embedding | Visual embedding | **BM25 exact matching** |
| Document Structure Utilization | None | None | **Hierarchical index (HDNC)** |

| Dimension | ColPali Family | VisRAG/M3DocRAG | **DVI (Ours)** |
|---|---|---|---|
| Information Fidelity | Indirect (embedding) | Direct (original image) | **Direct (original image)** |
| Engineering Document Validation | None | None | **Three datasets** |
| Bridge PageR@3 | 20.1% (measured) | — | **68.0%** |

The core novelty of DVI lies in two aspects: (1) **Zero VLM at preprocessing**—existing multimodal RAG methods all rely on vision models during preprocessing (to generate descriptions or embeddings), whereas DVI extracts textual metadata solely through rules; (2) **Leveraging document structure for retrieval**—existing approaches all employ flat vector stores, ignoring the hierarchical numbering systems of engineering documents, while DVI's HDNC algorithm automatically discovers and exploits this structure. It is worth emphasizing that we evaluated ColPali-v1.2 (ICLR 2025) on the Bridge dataset, where it achieved a PageR@3 of only 20.1%, even lower than naive text embedding (30.7%), demonstrating that visual embedding systematically fails on homogeneous engineering documents (see Section 4.2.1 for details).

# 3. Method: Deferred Visual Ingestion Framework

## 3.1 Framework Overview

The DVI framework decomposes document question answering into two stages (Figure 1):

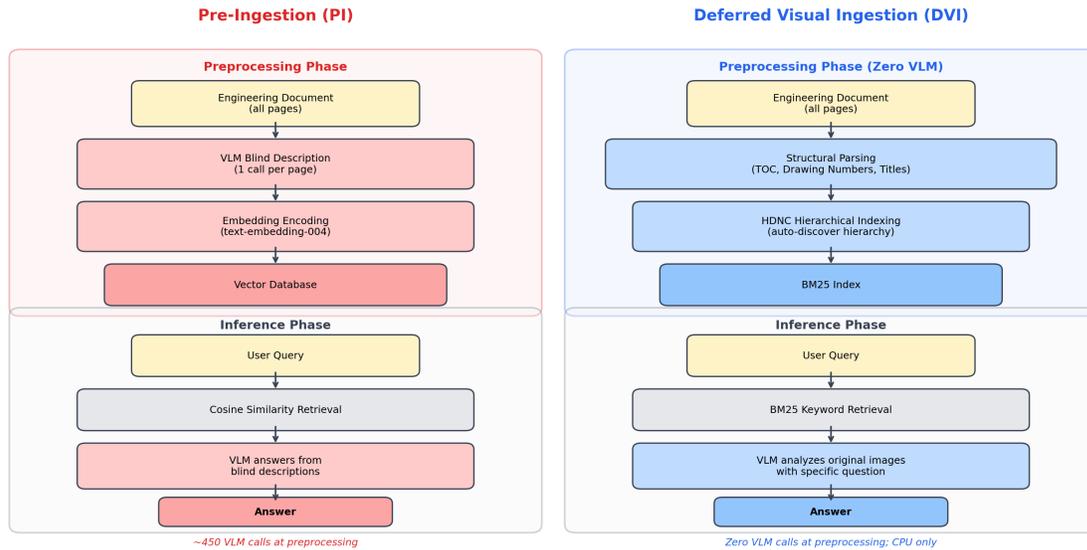

Figure 1: DVI vs. PI framework comparison — zero VLM calls during preprocessing vs. per-page VLM calls

**Preprocessing Stage (Zero VLM):** The document undergoes lightweight parsing to extract structured metadata (TOC, drawing numbers, titles). A hierarchical index is automatically constructed via the HDNC algorithm, and a BM25 search engine is built. The entire process runs on pure CPU computation without invoking any vision language model.

**Inference Stage (On-Demand VLM):** User query → BM25 retrieval locates top-k candidate pages → candidate pages are rendered as images → images and query are sent to a VLM for targeted visual analysis → answer is returned.

This design embodies the core principle of DVI: "**Index for locating, not understanding**." The goal of the preprocessing stage is not to "comprehend" the content of every drawing, but to establish sufficiently precise page mappings so that the inference stage can quickly locate the relevant 2–3 pages. The entire cost of visual understanding is deferred to the moment a user poses a specific question, and it occurs only for the few pages matched by retrieval.

The fundamental difference from Pre-Ingestion lies in when the vision model is invoked: PI calls the VLM on **every page** during preprocessing to generate blind descriptions (supply-side ingestion), whereas DVI calls the VLM only on **retrieval-matched pages** during inference for targeted analysis (demand-side ingestion). The former incurs VLM costs proportional to document size that are unavoidable; the latter incurs VLM costs proportional to actual query volume that can be continuously reduced through caching.

## 3.2 HDNC: Hierarchical Drawing Number Clustering for Automatic Indexing

HDNC (Hierarchical Drawing Number Clustering) is the core algorithm of the DVI preprocessing stage. It leverages the inherent structure of engineering drawing numbering systems to automatically construct a hierarchical index without any API calls.

### 3.2.1 Observation and Motivation

Engineering drawing numbers follow industry standards and exhibit systematic naming conventions within a given project. Taking the Bridge dataset in this paper as an example, over 500 drawings share the prefix `PROJID-GRP-PKG-ST-`, with trailing numeric codes encoding hierarchical information such as bridge identifier, drawing category, and sequence number:

```
PROJID-GRP-PKG-ST-BR-DR-101013   →   10|10|13   →   GA/Bridge-A
PROJID-GRP-PKG-ST-BR-DR-501521   →   50|15|21   →   Details/Pier/Pier-3
PROJID-GRP-PKG-ST-BR-DR-501616   →   50|16|16   →   Details/PT/Bridge-A
```

This encoding structure means that drawing numbers already contain rich hierarchical classification information. The goal of HDNC is to automatically discover and exploit this structure.

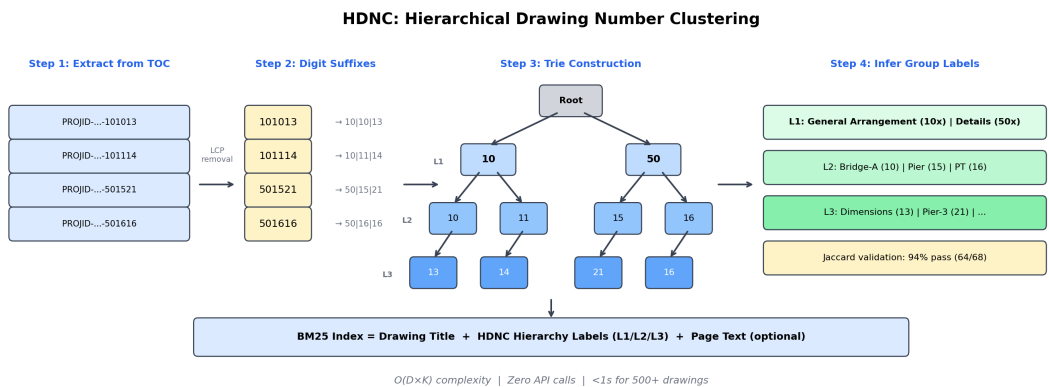

*Figure 2: HDNC algorithm pipeline — automatically discovering hierarchical structure from drawing numbers and constructing an index*

### 3.2.2 Algorithm Pipeline

HDNC consists of four steps:

**Step 1: TOC Parsing and Drawing Number Extraction.** A mapping list of drawing numbers and titles is extracted from the PDF's table of contents (TOC). Text is extracted using PyMuPDF, and drawing number patterns are identified via regular expressions. For documents without a TOC (e.g., Steel), drawing numbers can be extracted from the title block region through OCR or rule-based methods.

**Step 2: Longest Common Prefix Discovery.** The longest common prefix (LCP) is computed across all drawing numbers to identify the project-level fixed prefix. For Bridge, the LCP is `PROJID-GRP-PKG-ST-` (after stripping the drawing category suffixes `BR-DR-`/`ZZ-DR-`).

**Step 3: Numeric Suffix Trie Construction.** After removing the common prefix, the remaining numeric suffixes are segmented using different splitting strategies (e.g., every 2 digits, 3 digits, etc.), and a prefix tree (Trie) is constructed. The splitting strategy that yields the most balanced Trie is selected—a good split should produce multiple meaningful branches rather than one dominant branch with many leaves. The optimal split for Bridge is [3,1,1,1] digits (6-digit suffix split into 3+1+1+1), yielding 6 L1 categories, 14 L2 subcategories, and 68 L3 groups.

**Step 4: Group Label Inference and Validation.** For each Trie group, co-occurring words in the drawing titles within the group are extracted as group labels (e.g., "Pier Details", "Post Tensioning"). The Jaccard coefficient is used to validate grouping quality: the vocabulary overlap among drawings within the same group should be significantly higher than across groups. On Bridge, 94% of groups pass Jaccard validation (64/68 groups), indicating strong consistency between the drawing number hierarchy and the semantic hierarchy.

### 3.2.3 Index Document Construction

Based on the hierarchical structure produced by HDNC, a BM25 index document is constructed for each page:

```
Index Document = [Drawing Title] + [HDNC Hierarchical Labels] + [Page T
ext (optional)]
```

The hierarchical labels include the L1 category name, L2 subcategory name, and L3 group label, providing rich contextual information for each drawing. For example, the index document for the drawing "Bridge-A Pier-3 Dimension Details" would additionally contain hierarchical labels such as "Details Pier", enabling correct recall when searching for "pier details".

### 3.2.4 Complexity and Cost

The time complexity of HDNC is $O(D \times K)$, where $D$ is the number of drawings and $K$ is the maximum suffix length; typically $D < 1000$ and $K < 10$. The entire algorithm uses only PyMuPDF and the Python standard library, requiring **zero API calls and** **zero cost**. Over 500 drawings in Bridge complete in under 1 second on a single CPU core.

### 3.2.5 Scope of Applicability

HDNC requires documents with a systematic numbering scheme, which is nearly universal in the engineering drawing domain (ISO/IEC/ANSI standards all mandate structured numbering). For documents without a numbering scheme but with a TOC (e.g., the Steel catalog), DVI degrades to TOC parsing—extracting category names and page range mappings from the TOC to build a category-to-page-range index. For documents with neither numbering nor a TOC, DVI can fall back to BM25 full-text search alone.

## 3.3 BM25 Retrieval

DVI employs BM25 as its retrieval engine. For each user query, BM25 computes relevance scores over the index document collection and returns the top-k pages (k=3 in this paper).

The reason BM25 outperforms embedding-based vector retrieval on engineering documents is that engineering queries typically contain precise identifiers (drawing numbers, model numbers, specification values) that require exact matching rather than semantic approximation. For example, in the query "dimensions of Bridge-A Pier-3", "Bridge-A" and "Pier-3" are precise identifiers that BM25 can directly match to index documents containing these keywords. By contrast, embedding-based retrieval maps these identifiers into a semantic vector space where, among highly similar engineering documents (multiple bridges sharing similar structural descriptions), vectors for different pages cluster closely together, causing retrieval failure.

**Text Quality Adaptivity.** The incorporation of page text into index documents is optional and depends on text layer quality: - Vector PDFs (CAD exports): The text layer is complete and accurate; incorporating it substantially improves retrieval (Bridge +21.3pp). - High-quality OCR scanned documents: Selective incorporation. - Low-quality OCR scanned documents: No incorporation; only the structured index is used (validated on Steel).

## 3.4 Inference Stage: VLM Visual Answering

After retrieval returns the top-k pages, the system renders these pages as high-resolution images and sends them to the VLM along with the user's question. The VLM analyzes the original images with the specific question in mind and provides a targeted answer.

Compared to PI's blind descriptions, this "look at the image with a question in mind" approach has two fundamental advantages:

1. **Lossless Information.** The VLM directly analyzes the original images without any intermediate representation (blind descriptions, OCR, embedding), avoiding information loss.

2. **Question-Directed.** The VLM knows exactly what the user wants and can focus on the relevant visual details (e.g., specific dimension annotations, table rows, component identifiers), rather than generating comprehensive but generic descriptions.

### 3.5 Theoretical Comparison of DVI and PI

We compare DVI and PI along three core dimensions:

**Preprocessing Cost.** PI calls a VLM on every page to generate blind descriptions, incurring cost proportional to document size. Bridge (~450 pages) requires approximately 450 VLM calls. DVI's preprocessing consists of pure CPU operations at zero API cost.

**Information Fidelity.** During preprocessing, PI's VLM does not know what users will ask and can only generate generic blind descriptions, inevitably losing visual details. Critical information on engineering drawings—dimension annotations, terminal numbers, routing paths—is highly susceptible to omission. DVI sends original images to the VLM at inference time, preserving 100% information fidelity.

**Retrieval Reliability.** PI encodes blind descriptions as embedding vectors; on highly similar documents (drawings of the same type across multiple bridges), pages cluster tightly in vector space, causing retrieval failure. DVI uses BM25 to perform exact matching on drawing numbers, bridge names, and other identifiers, achieving far greater reliability than semantic vector retrieval on identifier-rich engineering documents. Experiments show that even in the most favorable scenario for embedding (Bridge vector PDFs with a complete text layer), BM25 retrieval substantially outperforms embedding retrieval in PageR@3 (68.0% vs. 30.7%).

## 4. Experiments

This section presents a systematic comparison of DVI and Pre-Ingestion (PI) across three datasets, covering both the retrieval layer and end-to-end QA.

### 4.1 Experimental Setup

#### 4.1.1 Datasets

We select three datasets spanning two document types (vector PDF and scanned document) and two sources (industrial data and public benchmark).

**Bridge: Bridge Engineering Drawings.** Approximately 450 pages of bridge design drawing packages (vector PDF), comprising over 20 bridges and more than 500 drawings. The documents possess a complete text layer and a standardized drawing number system (shared prefix `PROJID-GRP-PKG-ST-`, with numeric suffixes encoding bridge, category, and sheet sequence). The TOC (table of contents page) lists all drawing numbers and titles. This constitutes an ideal input for the HDNC algorithm.

**Steel: Steel Product Catalog.** An 83-page steel product catalog (full-page scanned document), containing 26 product categories (Universal Beams, Channels, Angles, etc.). Each PDF page contains two pages of the original book (spread scan). The document has a TOC page but no structured numbering system. OCR quality varies by page: the table of contents page has clear, readable text, while product table pages suffer from garbled OCR due to 90-degree rotation.

**CircuitVQA:** A public circuit diagram VQA benchmark comprising 1,077 circuit diagrams, 108 functional units (Units), and 9,315 question-answer pairs. Each Unit contains multiple sub-diagrams, with questions covering component identification, spatial relationships, and value reading. We use this dataset at the retrieval layer only.

### 4.1.2 QA Dataset Generation

The QA datasets for Bridge and Steel were generated with VLM assistance: each page was rendered as an image, and Gemini 3.1 Pro generated question-answer pairs based on the visual content of each page. To ensure question quality consistent with real engineering query scenarios, our prompt required that questions include specific locating information such as drawing numbers, product model numbers, or table names (e.g., "What is the length of Span-4 on drawing PROJID-GRP-PKG-ST-BR-DR-101013?"), rather than generic questions (e.g., "What is the span of the bridge?").

Bridge yielded 1,323 questions (dimension 756, value 258, identification 147, specification 134, count 28), with 82.9% containing locating information. Steel yielded 186 questions covering 62 pages and 26 categories.

**Limitation Disclaimer:** The QA datasets are synthetic, with questions and reference answers generated by a VLM. Although we required the inclusion of locating information to simulate real query styles, distributional differences between synthetic questions and real user queries may still exist. We further discuss this limitation in Section 5.3.

### 4.1.3 Compared Methods

**PI (Pre-Ingestion):** On Bridge, we employ a standard multimodal RAG pipeline—calling a VLM (Gemini 3.1 Pro) on each page to generate page descriptions, encoding them into vectors with text-embedding-004 for storage in a vector database, encoding queries with the same model at query time, performing cosine similarity retrieval, and sending the top-3 pages rendered as images to a VLM (Gemini 3.1 Pro) for answering. On Steel, since pages are scanned documents, PI uses Baidu OCR to extract text followed by BM25 retrieval (PI-OCR). On CircuitVQA, PI uses VLM blind description + embedding retrieval (PI-flat).

**DVI (Deferred Visual Ingestion):** Zero VLM calls during preprocessing. On Bridge, the HDNC algorithm automatically constructs a hierarchical index from the TOC; index documents consist of drawing titles + HDNC hierarchical labels + page text, with BM25 retrieving the top-3 pages that are then rendered as images and sent to a VLM for answering. On Steel, TOC parsing constructs a category-to-page mapping with BM25 retrieval. On CircuitVQA, Unit-level indexing + BM25 retrieval is used.

**Oracle:** The correct page images are directly sent to the VLM (Gemini 3.1 Pro) for answering, serving as a measure of the VLM capability ceiling.

### 4.1.4 Evaluation Metrics

**Retrieval Metrics:** PageR@k (the proportion of queries for which the target page appears in the top-k results), MRR (Mean Reciprocal Rank). For CircuitVQA, we additionally report ImgR@k (image-level) and UnitR@k (functional unit-level).

**End-to-End QA Metrics:** Accuracy (exact match rate between VLM answers and reference answers). For numerical questions, we apply numerical comparison with a tolerance of ±1%; for textual questions, we apply keyword matching.

### 4.1.5 Implementation Details

All VLM calls use Gemini 3.1 Pro. BM25 uses the rank_bm25 library with default parameters (k1=1.5, b=0.75). Embedding uses Google text-embedding-004 (768 dimensions). The HDNC algorithm is a pure Python implementation (PyMuPDF + standard library) with zero API calls. OCR for Steel uses Baidu OCR accurate_basic. The top-k is set to 3 for all experiments.

## 4.2 Retrieval Experiments

### 4.2.1 Bridge Drawing Retrieval

Table 1 reports retrieval results on the Bridge dataset for 1,323 questions.

**Table 1: Bridge Drawing Retrieval Results (1,323 questions)**

| Method | PageR@1 | PageR@3 | MRR | Preprocessing VLM Calls |
|---|---|---|---|---|
| ColPali-v1.2 [3] | 10.7% | 20.1% | 0.194 | ~450 (visual embedding) |
| PI-embedding | 16.9% | 30.7% | 0.229 | ~450 (blind description + embedding) |

| Method | PageR@1 | PageR@3 | MRR | Preprocessing VLM Calls |
| --- | --- | --- | --- | --- |
| DVI-title-only | 35.1% | 49.4% | 0.413 | 0 |
| DVI-HDNC-auto | 35.4% | 46.7% | 0.403 | 0 |
| **DVI-HDNC+text** | **55.4%** | **68.0%** | **0.610** | **0** |

DVI-HDNC+text substantially outperforms all embedding methods across every metric: PageR@3 68.0% vs. ColPali 20.1% (+47.9pp) vs. PI-embedding 30.7% (+37.3pp). Even the simplest DVI-title-only (BM25 indexing using only drawing titles) surpasses both embedding methods at 49.4% PageR@3.

**Systematic Failure of Embedding-Based Retrieval.** ColPali [3] (ICLR 2025) is a representative method for visual document retrieval, generating multi-vector embeddings (ColBERT-style MaxSim) directly from page screenshots using a vision-language model, and achieving strong performance on heterogeneous document benchmarks. However, on homogeneous engineering drawings, ColPali achieves only 20.1% PageR@3, even lower than naive PI-embedding (30.7%). This result demonstrates that the failure of embedding retrieval on highly similar documents is not due to an insufficiently powerful embedding model, but rather a structural deficiency of the retrieval paradigm itself: bridge drawings from the same project are highly similar both visually and semantically, and neither single-vector (text-embedding) nor multi-vector (ColPali) representations can effectively discriminate in the vector space. Head-to-head analysis reveals that ColPali and PI-embedding exhibit different hit patterns (ColPali uniquely wins 146 questions vs. PI uniquely wins 286 questions), yet both fail systematically. In contrast, BM25's exact keyword matching can precisely locate pages using identifiers such as drawing numbers and bridge names.

### 4.2.2 Steel Catalog Retrieval

Table 2 reports retrieval results on the Steel dataset for 186 questions.

**Table 2: Steel Catalog Retrieval Results (186 questions)**

| Method | PageR@1 | PageR@3 | MRR |
| --- | --- | --- | --- |
| **DVI-TOC** | **31.7%** | **65.6%** | **0.462** |
| DVI-TOC+OCR | 19.4% | 24.7% | 0.215 |
| PI-OCR | 16.1% | 23.1% | 0.189 |

DVI-TOC substantially outperforms PI-OCR: PageR@3 65.6% vs. 23.1% (+42.5pp). Notably, DVI-TOC uses only category names and page number mappings parsed from the TOC page, with no dependence on page content text whatsoever.

**Adverse Effect of OCR Text Fusion.** DVI-TOC+OCR augments DVI-TOC with OCR-extracted page text, yet performance drops sharply (65.6% → 24.7%, −40.9pp). The reasons are twofold: (1) OCR quality on the scanned document is extremely poor (table pages suffer from garbled text due to 90-degree rotation), introducing substantial noise; (2) the TOC page itself contains all category names, and during BM25 retrieval the TOC page receives the highest score (in 131 of 186 questions, the TOC page enters the top-3), crowding out the actual target pages. This stands in stark contrast to the positive effect of text fusion on Bridge (discussed in Section 4.4).

### 4.2.3 CircuitVQA Retrieval

Table 3 reports retrieval results on the CircuitVQA dataset for 9,315 questions.

**Table 3: CircuitVQA Retrieval Results (9,315 questions)**

| Method | ImgR@1 | ImgR@3 | UnitR@1 | UnitR@3 |
| --- | --- | --- | --- | --- |
| PI-flat | 0.3% | 0.7% | 3.8% | 25.8% |
| **DVI-BM25** | **10.9%** | **31.2%** | **98.4%** | **99.4%** |

DVI achieves near-perfect Unit-level retrieval (UnitR@3 99.4%), while PI-flat reaches only 25.8%. At the image level, DVI also leads by a wide margin (ImgR@3 31.2% vs. 0.7%). PI's embedding retrieval almost completely fails on 1,077 highly similar circuit sub-diagrams—sub-diagrams from different Units are highly similar both visually and semantically, and the vector space cannot effectively discriminate among them.

### 4.2.4 Cross-Dataset Analysis

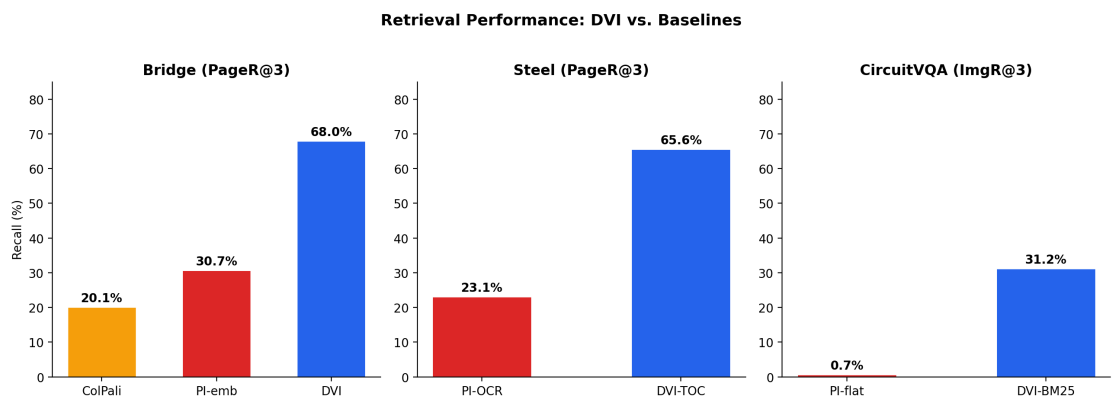

*Figure 3: Retrieval performance comparison across three datasets — DVI substantially outperforms across all datasets*

The retrieval experiments across all three datasets yield consistent findings:

1. **DVI substantially outperforms PI on all datasets.** Bridge +37.3pp, Steel +42.5pp, CircuitVQA +30.5pp (ImgR@3).

2. **Embedding retrieval systematically fails on highly similar documents.** ColPali (ICLR 2025 SOTA) achieves only 20.1% PageR@3 on Bridge, even lower than naive text embedding (30.7%). Engineering drawings are highly similar both visually and textually, and the vector space cannot effectively discriminate regardless of whether single-vector or multi-vector representations are used.

3. **DVI's zero preprocessing cost is a genuine advantage.** PI-embedding requires calling a VLM on every page (approximately 450 calls for Bridge) plus encoding into vectors, whereas DVI needs only to parse the TOC and text layer, with zero API calls throughout.

## 4.3 End-to-End QA Experiments

### 4.3.1 Main Results

Table 4 reports end-to-end QA results for Bridge and Steel, concatenating retrieval and VLM answering into a complete pipeline.

**Table 4: End-to-End QA Results**

| Dataset | DVI E2E | PI E2E | Oracle | DVI Advantage |
|---|---|---|---|---|
| Bridge (1,323 questions) | **65.6%** | 24.3% | 93.0% | **+41.3pp** |
| Steel (186 questions) | **30.6%** | 16.1% | 83.3% | **+14.5pp** |

On Bridge, DVI leads PI by a striking +41.3pp margin. Head-to-head analysis reveals: DVI uniquely wins 631 questions vs. PI uniquely wins 84 questions (7.5:1), with 237 questions answered correctly by both and 371 questions missed by both. DVI's advantage is overwhelming.

On Steel, DVI leads by +14.5pp. Head-to-head: DVI uniquely wins 48 questions vs. PI uniquely wins 21 questions (2.3:1). Steel's lower absolute accuracy (DVI 30.6%, PI 16.1%) is primarily attributable to two factors: (1) scanned document image quality limits VLM reading capability (Oracle at only 83.3%, lower than Bridge's 93.0%); (2) cross-page questions (when a product category spans 3+ pages, top-3 retrieval can only hit a subset of pages).

### 4.3.2 Retrieval-to-QA Propagation Analysis

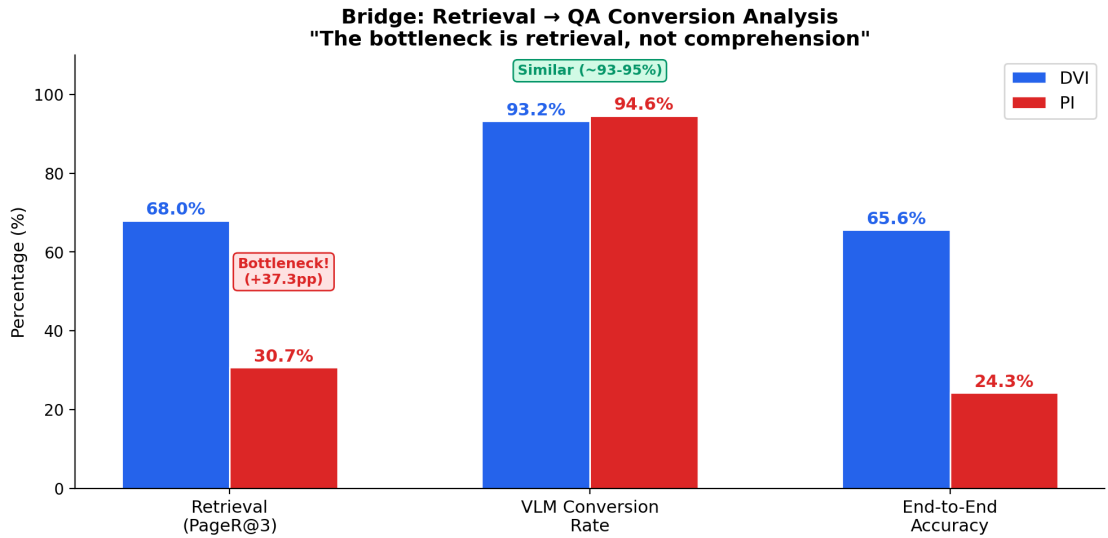

*Figure 4: Bridge retrieval-to-QA conversion analysis — the bottleneck lies at the retrieval stage, not at VLM comprehension*

The VLM conversion rate given the correct page is remarkably high: 93.2% for DVI and 94.6% for PI on Bridge. **The VLM conversion rates for both methods are nearly identical; the bottleneck lies entirely at the retrieval stage.** DVI's end-to-end advantage derives directly from its retrieval advantage (Bridge PageR@3 68.0% vs. 30.7%), and VLM capability itself is not a differentiating factor.

This finding carries important system design implications: in visually dense document QA, investing in retrieval improvement is more effective than investing in VLM comprehension capability. DVI's design principle of "indexing solely for localization" is grounded precisely in this insight.

### 4.3.3 Analysis by Question Type (Bridge)

Table 5 breaks down Bridge end-to-end results by question type.

**Table 5: Bridge End-to-End QA Breakdown by Question Type**

| Question Type | Count | DVI Accuracy | PI Accuracy | Oracle | DVI Advantage |
|---|---|---|---|---|---|
| dimension | 756 | 57.5% | 20.8% | 91.7% | +36.8pp |
| value | 258 | 88.0% | 26.0% | 96.1% | +62.0pp |
| identification | 147 | 66.0% | 40.1% | 96.6% | +25.9pp |
| specification | 134 | 63.4% | 23.1% | 88.8% | +40.3pp |
| count | 28 | 85.7% | 25.0% | 100.0% | +60.7pp |

DVI substantially outperforms PI across all question types. The value and count types exhibit the largest DVI advantages (+62.0pp and +60.7pp); these question types require precise localization of pages containing specific numerical values, a scenario where PI's embedding retrieval is particularly weak. The identification type shows the smallest DVI advantage (+25.9pp) but remains significant—PI performs relatively better on identification questions, likely because VLM blind descriptions provide more complete coverage of component names.

## 4.4 Ablation Studies

### 4.4.1 HDNC Index Ablation

Table 6 compares retrieval performance of three indexing strategies on an 80-question subset of Bridge.

**Table 6: HDNC Index Ablation (Bridge, 80-question subset)**

| Method | PageR@1 | PageR@3 | MRR | Description |
| --- | --- | --- | --- | --- |
| DVI-title-only | 26.2% | 47.5% | 0.356 | Drawing titles only |
| **DVI-HDNC-auto** | **42.5%** | **75.0%** | **0.560** | HDNC automatic hierarchical labels |
| DVI-manual | 53.8% | 68.8% | 0.608 | VLM-extracted structured metadata |

HDNC automatic indexing yields a substantial improvement: from title-only to HDNC-auto, PageR@3 increases from 47.5% to 75.0% (+27.5pp), and MRR from 0.356 to 0.560 (+57%). This improvement derives entirely from zero-cost algorithmic inference—HDNC automatically discovers hierarchical structure from drawing number prefixes (6 L1 categories, 14 L2 subcategories, 68 L3 groups) and generates hierarchical membership labels for each drawing.

Notably, HDNC-auto even surpasses DVI-manual in PageR@3 (75.0% vs. 68.8%, +6.2pp), where the latter uses VLM-extracted structured metadata from drawing title blocks. This indicates that HDNC's hierarchical labels provide complementary information—drawings within the same category share hierarchical paths, creating a "neighborhood effect" whereby even when a query does not precisely match the target drawing, adjacent drawings in the same group are recalled. Manual performs better on PageR@1 and MRR (precise metadata aids top-1 localization), suggesting the two approaches are complementary rather than substitutive.

### 4.4.2 Text Quality Adaptiveness

Text fusion produces diametrically opposite effects across different document types:

**Table 7: Text Fusion Effect Comparison**

| Dataset | PageR@3 w/o Text | PageR@3 w/ Text | Difference | Document Type |
|---|---|---|---|---|
| Bridge | 46.7% (HDNC-auto) | 68.0% (HDNC+text) | **+21.3pp** | Vector PDF |
| Steel | 65.6% (TOC) | 24.7% (TOC+OCR) | **-40.9pp** | Scanned document |

Bridge is a vector PDF whose text layer is directly output by CAD software—clean and accurate data that substantially improves retrieval upon fusion. Steel is a scanned document with extremely poor OCR text quality (rotated tables produce garbled text), and fusion introduces noise that proves detrimental.

**The SPECIAL STEELS Counterexample in Steel.** All 15 questions in the SPECIAL STEELS category of Steel are missed by DVI-TOC (0/15), yet DVI-TOC+OCR and PI-OCR hit 13/15 and 13/15 respectively. This category happens to have good OCR quality, making text retrieval effective. This counterexample supports the necessity of an adaptive strategy: **the decision of whether to fuse text information should be made dynamically based on the text quality of each document or even each page.**

**Adaptive Strategy Framework:** - Vector PDF (complete text layer) → Fuse page text; BM25 searches both index and text - High-quality scanned document (high OCR confidence) → Selectively fuse OCR text - Low-quality scanned document (garbled OCR) → Use only structured index (TOC/HDNC); do not fuse text

## 4.5 Summary of Experiments

The experiments across three datasets consistently demonstrate the core advantages of DVI:

1. **DVI comprehensively outperforms PI in both retrieval and end-to-end QA at zero VLM preprocessing cost.** Bridge E2E +41.3pp, Steel E2E +14.5pp, retrieval layer +30pp or more across all three datasets.

2. **The bottleneck lies at the retrieval stage, not at VLM comprehension.** The VLM conversion rate given the correct page exceeds 93%; improving retrieval is the key to enhancing end-to-end performance.

3. **HDNC zero-cost automatic indexing is effective.** Automatically inferring hierarchical structure solely from drawing number prefixes yields a +27.5pp retrieval improvement and even surpasses VLM-extracted manual metadata in recall.

4. **Text quality adaptiveness is critical for real-world deployment.** Text fusion is beneficial for vector PDFs but harmful for scanned documents;

indexing strategies must be dynamically adjusted according to document type.

# 5. Discussion

## 5.1 Text Quality Adaptive Strategy

Our experiments reveal an important finding: text fusion produces diametrically opposite effects across different document types. For Bridge (vector PDF), fusing page text yields a +21.3pp retrieval improvement, whereas for Steel (scanned document), fusing OCR text causes a −40.9pp retrieval degradation. This discrepancy stems from fundamental differences in text layer quality: vector PDF text is directly output by CAD software, making it accurate and complete; OCR from scanned documents is severely affected by image quality, rotation angles, and table complexity, resulting in substantial noise.

The counterexample of the SPECIAL STEELS category in Steel further reveals the fine-grained nature of this issue: OCR quality can vary dramatically across different pages within the same document. This category has relatively good OCR quality, and fusing OCR text significantly improved retrieval (0/15 → 13/15). This indicates that the optimal strategy is not a document-level binary choice (fuse or not fuse), but rather a page-level adaptive decision.

**Unified framework:** We propose a text quality adaptive indexing strategy: - Detect document type (vector PDF / scanned document) - For scanned documents, assess per-page OCR quality (based on confidence scores, character recognition rates, etc.) - Fuse OCR text for high-quality pages; use only structural indexing for low-quality pages

The concrete implementation and validation of per-page OCR quality assessment are left for future work.

## 5.2 Applicability Boundaries of HDNC

The HDNC algorithm relies on systematic naming conventions for engineering drawing numbers. DVI adopts different indexing strategies for three categories of scenarios:

1. **Document sets with structured numbering (e.g., Bridge engineering drawings):** HDNC automatically discovers hierarchical structure and constructs multi-level taxonomic indices. This is the optimal scenario for DVI, and validation results show that automatic indexing achieves recall even higher than manual VLM-extracted metadata.

2. **Documents with TOC but no systematic numbering (e.g., Steel product catalog):** Category-to-page mappings are extracted from the TOC to construct a flat index. Although hierarchical information is limited, the category-level localization provided by the TOC (CatR@3 85.5%) still far surpasses PI.

3. **Documents without structural information:** Only BM25 full-text search is used. DVI degrades to pure text retrieval + inference-time VLM visual analysis, still preserving the advantage of information losslessness, but retrieval capability is constrained by text layer quality.

The key assumption of HDNC—that drawing numbers within the same project follow consistent naming conventions—holds nearly universally in engineering practice, as industry standards such as ISO, IEC, and ANSI all mandate structured numbering. The 94% Jaccard validation pass rate (64/68 groups) on the Bridge dataset provides empirical support for this assumption.

## 5.3 Limitations of DVI

**Synthetic QA datasets.** The QA datasets for Bridge and Steel are generated by VLMs rather than derived from real user queries. Although we prompted for realistic questions containing localization information (82.9% include localization information), synthetic questions may still exhibit distributional differences from actual engineer queries. The substantial discrepancy between v1 generic questions (only 9.1% with localization information) and v2 results demonstrates that question distribution significantly affects experimental conclusions. Validation on real customer queries is needed in the future.

**Inference-time VLM latency.** DVI requires invoking a VLM to analyze raw images at inference time, introducing an additional latency of 20–40 seconds per query. For time-sensitive application scenarios (e.g., engineers consulting drawings on-site), this latency may be unacceptable. Progressive Knowledge Accumulation (PKA) can partially mitigate this issue—as usage accumulates, queries for cached pages can return with zero latency.

**Dataset scale and domain coverage.** The industrial data covers only two domains (bridge engineering, steel products) across two documents. Although CircuitVQA provides cross-validation on a public benchmark, the generalizability of DVI across additional industries (architecture, mechanical, electronics, chemical) remains to be verified. This framework has been demonstrated on low-voltage switchgear engineering drawings (over 100 pages, over 20 drawings) from an electrical equipment manufacturer, and the client is in the process of signing a service agreement based on the demonstration results. Due to non-disclosure agreement restrictions, specific data are not disclosed in this paper.

**Limitations of evaluation metrics.** The current Accuracy metric uses exact matching, which is highly sensitive to OCR misreadings (e.g., VLM misreading "MMD" as "MWD," causing an otherwise correct answer to be judged incorrect). A more robust semantic matching evaluation may more accurately reflect the system's practical usefulness.

## 5.4 Future Work

**Validation on real customer queries.** Collect real query logs from engineers in actual deployment environments and cross-validate against conclusions drawn from synthetic QA datasets.

**Adaptive text fusion strategy.** Implement page-level OCR quality assessment to automatically determine whether to fuse text information into the BM25 index for each page.

**Progressive Knowledge Accumulation (PKA).** VLM analysis results at inference time are automatically cached at the page × query_type granularity, enabling zero VLM cost for subsequent queries of the same type. As usage accumulates, VLM invocation rates continuously decrease, with knowledge progressively accumulating driven by real demand.

**Engineering drawing QA benchmark.** This domain lacks a publicly available standardized evaluation set. Collecting engineering documents across multiple industries and formats and constructing a standardized QA benchmark would help advance systematic research.

# 6. Conclusion

This paper presents Deferred Visual Ingestion (DVI), a framework that defers the visual understanding of multimodal document QA from the preprocessing stage to the inference stage. The core principle of DVI—"index only to locate, not to comprehend"—employs the HDNC algorithm to automatically construct hierarchical indices and BM25 retrieval for page localization, sending raw images along with specific questions to a VLM for on-demand analysis.

Large-scale experiments across three datasets validate that DVI comprehensively outperforms traditional Pre-Ingestion approaches at zero VLM preprocessing cost: Bridge engineering drawings (1,323 questions) end-to-end QA accuracy 65.6% vs. 24.3% (+41.3pp), Steel product catalog (186 questions) 30.6% vs. 16.1% (+14.5pp), and CircuitVQA (9,315 questions) retrieval ImgR@3 31.2% vs. 0.7% (+30.5pp). VLM conversion rate analysis demonstrates that the bottleneck lies entirely on the retrieval side (conversion rate >93% once the correct page is obtained), validating the design insight that "investing in retrieval outweighs investing in comprehension."

HDNC zero-cost automatic indexing discovers hierarchical structure from drawing numbers, yielding a +27.5pp retrieval improvement, with recall even surpassing manually VLM-extracted metadata. The text quality adaptive strategy finding (vector

PDF +21.3pp vs. scanned document −40.9pp) provides critical guidance for practical deployment.

The philosophy of "index only to locate, not to comprehend" demonstrates strong practical effectiveness in engineering document RAG. Just as lazy evaluation computes only the values that are actually needed, DVI analyzes only the drawings that are actually queried—not all information needs to be "understood" in advance; reading with a question in mind is the best way to comprehend.